\documentclass[3p,preprint,twocolumn,times]{elsarticle}

\usepackage{prletters}
\usepackage{threeparttable}
\usepackage{amsmath}
\usepackage{makecell}
\usepackage{graphicx}
\usepackage{multirow}
\usepackage{algorithm}
\usepackage{algorithmic}
\usepackage{color}
\usepackage{booktabs}
\usepackage{amssymb}
\biboptions{sort&compress}
\usepackage[figuresright]{rotating}

\begin{document}

\begin{frontmatter}
\title{Learnable Chamfer Distance for Point Cloud Reconstruction}

\author[label2,label1]{Tianxin Huang}
\author[label1]{Qingyao Liu}
\author[label1]{Xiangrui Zhao}
\author[label1]{Jun Chen}
\author[label1]{Yong Liu\corref{cor1}}
\cortext[cor1]{denotes Corresponding author\\Email: T.Huang~(21725129@zju.edu.cn),~Q.Liu~(qingyaoliu@zju.edu.cn),\\X.Zhao~(xiangruizhao@zju.edu.cn),~J.Chen~(junc@zju.edu.cn),\\Y.Liu~(yongliu@iipc.zju.edu.cn)}
%\cortext[cor2]{Email:21725129@zju.edu.cn(T.Huang),}
\address[label2]{National University of Singapore, Singapore}
\address[label1]{Zhejiang University, Hangzhou, China}
%\address[label1]{Laboratory of Advanced Perception on Robotics and Intelligent Learning, College of Control Science and Engineering, Zhejiang University, Hangzhou, China}
%\address[label2]{Beijing Institute of Mechanical and Electrical Engineering, Beijing, China}

%\address{}
%\thanks{Email:21725129@zju.edu.cn(T.Huang)}
\begin{abstract}
	%For the related tasks of point cloud reconstruction, reconstruction losses are necessary to train the networks with the measurements of shape differences between ground truths and reconstructed point clouds. 
	As point clouds are 3D signals with permutation invariance, most existing works train their reconstruction networks by measuring shape differences with the average point-to-point distance between point clouds matched with predefined rules. 
	However, the static matching rules may deviate from actual shape differences.
	Although some works propose dynamically-updated learnable structures to replace matching rules, they need more iterations to converge well.
	In this work, we propose a simple but effective reconstruction loss, named Learnable Chamfer Distance (LCD) by dynamically paying attention to matching distances with different weight distributions controlled with a group of learnable networks.
	By training with adversarial strategy, LCD learns to search defects in reconstructed results and overcomes the weaknesses of static matching rules, while the performances at low iterations can also be guaranteed by the basic matching algorithm.
	Experiments on multiple reconstruction networks confirm that LCD can help achieve better reconstruction performances and extract more representative representations with faster convergence and comparable training efficiency. The source codes are provided in https://github.com/Tianxinhuang/LCDNet.git.
\end{abstract}
%\vskip -0.3in
\begin{keyword}
Keywords: 3D point cloud processing, reconstruction loss, adversarial strategy
\end{keyword}
\end{frontmatter}
%\maketitle

\section{Introduction}

Point cloud is one signal describing the 3D shape, which is widely-used due to its convenient acquisition from 3D sensors such as RGB-D camera or LiDAR. Different from regular 1-D signals or 2-D images, point clouds are permutation-invariant, which means changing specific permutations of points does not change described shapes. In other words, the permutations of points do not include any useful information.
In this condition, commonly-used mean squared errors~(MSE) cannot be directly applied to point cloud reconstruction. To train a point cloud reconstruction network, most existing works use the Chamfer Distance (CD) or Earth Mover's Distance (EMD)~\cite{fan2017point} as training losses.
They match points with predefined rules and measure shape differences between input point clouds and reconstructed results by average point-to-point distance.
However, the losses based on manually-defined matching rules are \emph{static}, which means the optimization goals are fixed and unchanged for all data during training. They may deviate the actual shape differences and make the reconstruction fall into local minimums with inferior reconstructed results but low reconstruction losses. 
Although some works~\cite{huang2020pf,wang2020cascaded,li2019pu} introduce GAN discriminators~\cite{NIPS2014_5ca3e9b1} to improve the reconstruction performance, they simply add the discriminator constraints to CD or EMD. Their improvements are limited as the discriminators only provide slight corrections to unchanged CD or EMD as shown in \cite{huang2022learning}.
%which can only produce slight improvements as their optimization still mainly reply on unchanged CD or EMD.
%whose optimization still mainly rely on matching rules and hard to get rid of their limitations.
PCLoss~\cite{huang2022learning} replaces the matching-based losses with distances between comparison matrices extracted with \emph{dynamic}-updated learnable structures, which totally avoids the adoption of static matching rules and learns to use changing measurements to measure the shape differences. It learns to search the shape defects by adversarial process, which has better performances due to the removal of predefined rules. But the totally learnable structures perform relatively inferior at the beginning of training process because it needs iterations to learn to find the defects.
%has relatively poor performances at the beginning of training process.

\begin{figure}[t]
	%\vskip -0.3in
	\begin{center}
		\scalebox{1.0}{
			\centerline{\includegraphics[width=\linewidth]{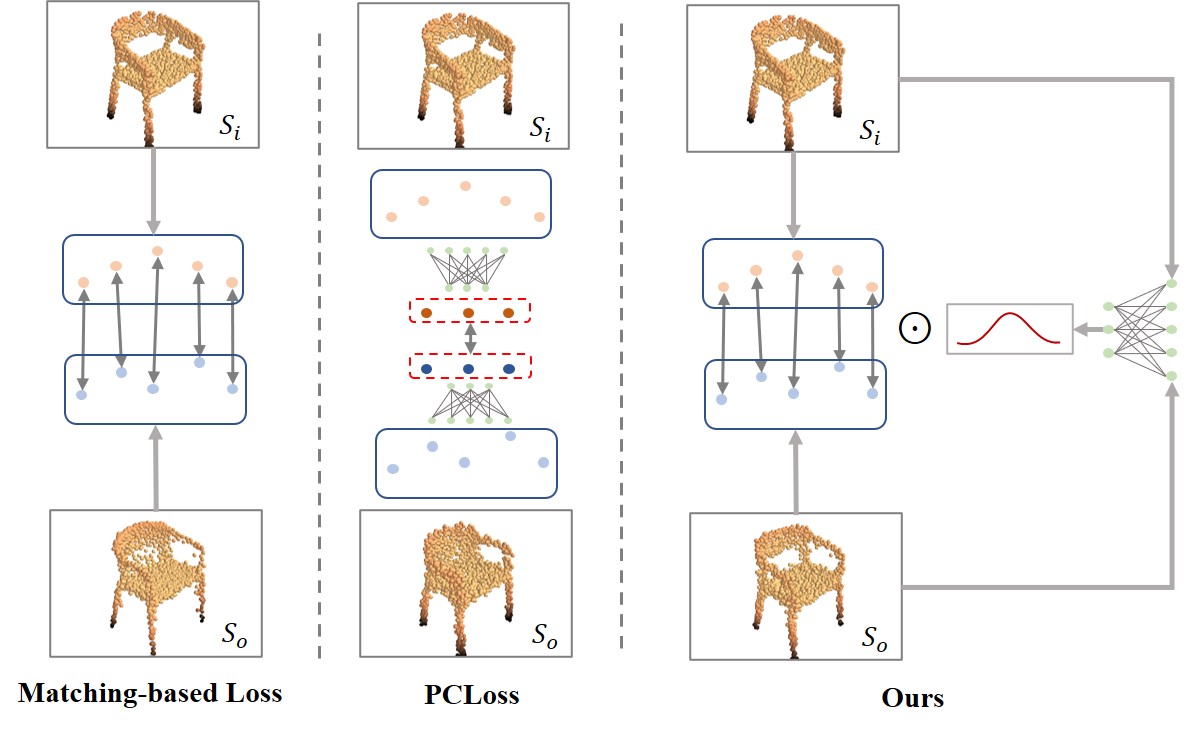}}
		}
		%\vskip -0.1in
		\caption{Differences between our method and existing reconstruction losses. Unlike the totally learning-based loss evaluation by extracting comparison matrices in PCLoss, our method uses networks to dynamically predict weight distributions for matching distances.}
		%\vskip -0.2in
		\label{pic:intro}
	\end{center}
\end{figure}

Considering the problems mentioned above, we propose a simple but effective learnable point cloud reconstruction loss, named Learnable Chamfer Distance (LCD) by designing a reasonable combination of \emph{dynamic} learning-based strategy and \emph{static} matching-based loss evaluation. 
The differences between LCD and existing methods are presented in Fig.~\ref{pic:intro}. Unlike the totally learning-based design in PCLoss~\cite{huang2022learning}, LCD learns to predict weight distributions for matching distances of different points.
During training, LCD is optimized by turns with the reconstruction network through an adversarial strategy to search regions with more shape defects, where the weight distributions are dynamically adjusted to pay more attention to matching distances of different regions.
Benefited from the adoption of dynamic learning-based strategy, LCD can achieve outstanding performances for the training of reconstruction networks, while the static matching-based evaluation can provide an initialization prior for the optimization and ensure that LCD has better performances than totally learning-based PCLoss~\cite{huang2022learning} at the beginning of training process.
%static matching-based evaluation, the optimization can go through an initial 
%the reconstruction loss is often calculated by matching points with predefined rules and  the average point-to-point distance 

Our contributions can be summarized as 
\begin{itemize}
	\item We propose Learnable Chamfer Distance (LCD), which can learn to search shape defects by dynamically predicting weight distributions for matching distances;
	\item Benefited from the reasonable combination of learning-based strategy and matching-based evaluation, LCD has faster convergence than existing learning-based losses;
	\item Experiments on multiple point clouds reconstruction networks demonstrate that LCD can help the reconstruction networks achieve better reconstruction performances and extract more representative representations.
	%point cloud reconstruction, unsupervised classification, and point cloud completion demonstrate that CALoss can help the task network improve reconstruction performances and learn more representative representations with higher training efficiency.% than existing learning-based methods.
\end{itemize}
%As the reconstruction loss is necessary for all reconstruction-related applications, such as point cloud representation learning, it is valuable to conduct corresponding researches to further stimulate the network potential. 
\begin{figure*}[t]
	%\vskip -0.3in
	\begin{center}
		\scalebox{1.0}{
			\centerline{\includegraphics[width=\linewidth]{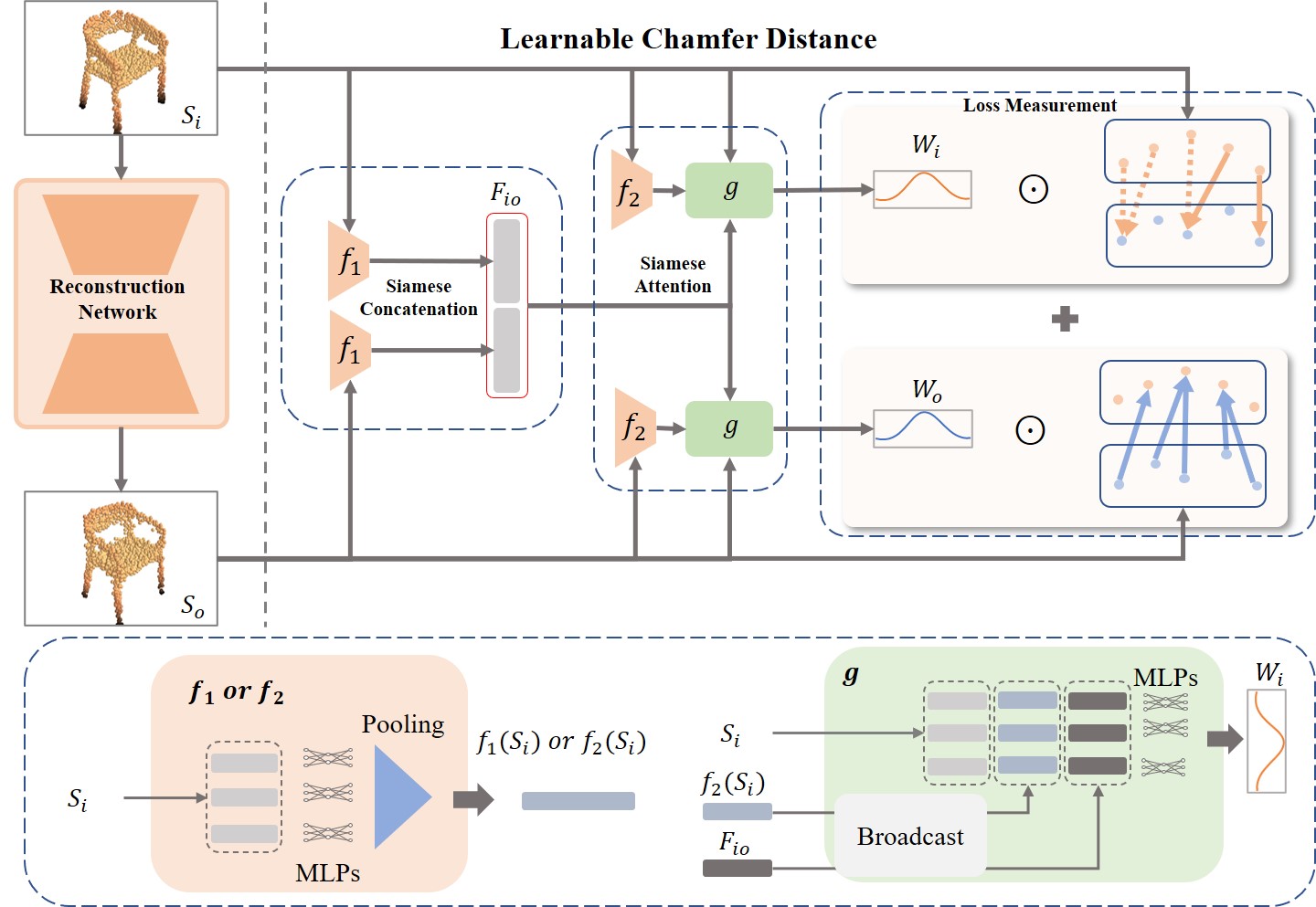}}
		}
		%\vskip -0.1in
		\caption{The pipeline of Learnable Chamfer Distance. $f_1$ and $f_2$ are constructed following PointNet~\cite{qi2017pointnet} framework to extract global features, while $g$ merges multiple features to predict weight distributions $W_i$ and $W_o$ for matching distances. The input point cloud $S_i$ and reconstructed point cloud $S_o$ are introduced to Siamese Concatenation module to extract global shape representation $F_{io}$, which is injected into Siamese Concatenation module to predict weight distributions $W_i$ and $W_o$ for matching distances between $S_i$ and $S_o$.}
		\label{pic:pipeline}
		%\vskip -0.9in
	\end{center}
\end{figure*}
\section{Related Works}
\subsection{Point Cloud Reconstruction}
Point cloud reconstruction aims to design networks, e.g. auto-encoders, to reconstruct point clouds through the representation extracted from the input point clouds. It can be adopted to related tasks like completing~\cite{wang2020cascaded,yuan2018pcn,wang2022mutual,huang2021rfnet,liu2020morphing} or sampling~\cite{lang2020samplenet,dovrat2019learning,zhu2022curvature} point clouds, while the extracted intermediate representations can be used for the unsupervised classification~\cite{yang2018foldingnet,achlioptas2018learning,huang2022resolution}.
Following PCLoss~\cite{huang2022learning}, the basic point cloud reconstruction network is often organized with encoders to extract representations from point clouds, and decoders to generate point clouds from the intermediate representations. The commonly-used encoders include PointNet~\cite{qi2017pointnet}, PointNet++~\cite{qi2017pointnet++}, and DGCNN~\cite{wang2019dynamic}, while decoders often come from fully connected networks proposed in AE~\cite{achlioptas2018learning} and FoldingNet proposed in \cite{yang2018foldingnet}.
In this work, we follow PCLoss~\cite{huang2022learning} to construct multiple reconstruction networks to evaluate performances of different losses.
%use multiple combinations of these encoders and decoders as the reconstruction networks to evaluate the performances of different training losses.

\subsection{Reconstruction Loss Design}
\label{matchloss}
Most existing point cloud reconstruction-related tasks rely on the Chamfer Distance (CD)~\cite{yuan2018pcn} and Earth Mover's Distance (EMD)~\cite{fan2017point}, which evaluate the reconstruction losses based on the average point-to-point distance between matched input and reconstructed point clouds. However, the predefined matching rules are static, which may cause the training processes fall into local minimums due to the deviation of predefined rules.
In this condition, many researchers attempt to introduce learning-based strategy to improve the constraining performances. Most researchers~\cite{huang2020pf,wang2020cascaded,li2019pu} design GAN discriminators~\cite{NIPS2014_5ca3e9b1} for extra supervisions. However, these works simply add the discriminator constraints to basic CD or EMD losses. Their optimizations still mainly reply on the matching-based CD or EMD, where the discriminators can only provide slight corrections. Therefore, these methods often have limited improvements.
Although DCD~\cite{wu2021density} fixes the matching rules in CD by considering the density distribution of reconstructed points, it is still limited by the static evaluation for reconstruction losses.

%Their performances are still limited by the static predefined matching rules. 
PCLoss~\cite{huang2022learning} replaces the usage of static matching rules with a dynamic learning process. It learns to extract comparison matrices from point clouds with differentiable structures and measure shape differences with distances between comparison matrices. 
%PCLoss is trained with adversarial strategy dynamically to consistently search the shape differences.
But the totally learning-based training process in PCLoss makes it need more iterations to converge well, while the relatively complex structures bring low training efficiency.
In this work, we explore to organically combine the learning-based strategy with matching rules by learning to pay attention to different matching connections.
Benefited from the learning-based strategy, our method has better performances than matching-based methods, while the adoption of static rules ensures it has faster convergence than totally learning-based methods like PCLoss.
%In this work, we explore to organically combine the learning-based strategy with matching rules by learning to pay attention to different matching connections.
%Benefited from the learning-based strategy, our method has better performances than matching-based methods, while the adoption of matching-based rules ensure it has faster convergence than fully learning-based methods like PCLoss.

\section{Methodology}
In this work, we propose a new method named Learnable Chamfer Distance (LCD) to evaluate the reconstruction loss by measuring the average point-to-point distance weighted with dynamically updated distributions. In this work, we use the static matching rules in CD~\cite{yuan2018pcn} to calculate the matching distances due to its high efficiency.
%The matching rule in CD is used due to its high efficiency.
The structure of LCD is presented in Sec.~\ref{CAM}. 
The training process of the reconstruction network with LCD is presented in Sec.~\ref{training}.
\subsection{The Structure of Learnable Chamfer Distance}
\label{CAM}
We propose a series of learnable structures to dynamically predict the weight distributions for the matching distances of different points.
As shown in Fig.~\ref{pic:pipeline}, the weight distributions $W_i$ and $W_o$ are predicted with Siamese Concatenation block (\textbf{SiaCon}) and Siamese Attention block (\textbf{SiaAtt}). \textbf{SiaAtt} predicts weight distributions for the matching distances, while \textbf{SiaCon} extracts global shape representations from both input and reconstructed point clouds and injects them to \textbf{SiaAtt}. 
\textcolor{black}{Specifically, in \textbf{SiaCon} block, two parameter-shared $f_1(\cdot)$ are used to extract global features from input point cloud $S_i$ and reconstructed result $S_o$. They are concatenated to construct a overall perception for the shapes $S_i$ and $S_o$. In \textbf{SiaAtt} block, two global features extracted by $f_2$ include independent shape information of $S_i$ and $S_o$, respectively. This information is fused with overall perception of two models to predict a weight for each coordinates with MLP in $g(\cdot)$.}
\begin{figure*}[t]
	%\vskip -0.3in
	\begin{center}
		\scalebox{1.0}{
			\centerline{\includegraphics[width=\linewidth]{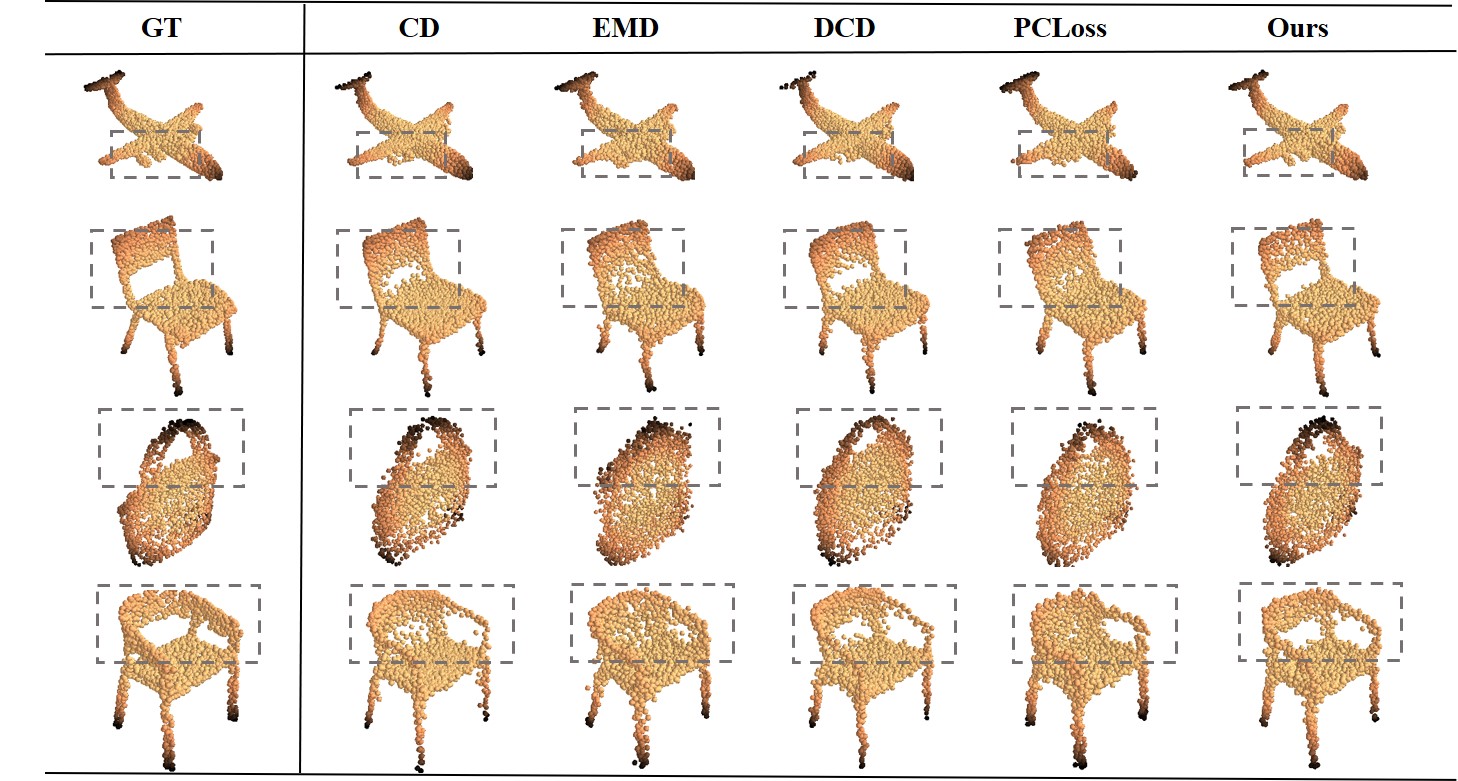}}
		}
		%\vskip -0.1in
		\caption{Qualitative comparison between LCD and other methods based on AE~\cite{achlioptas2018learning} following PCLoss~\cite{huang2022learning}. Our method can help create clearer details as shown in the circled regions.}
		\label{pic:quali}
		%\vskip -0.6in	
	\end{center}
\end{figure*}

Let $S_i$ and $S_o$ be the input point clouds and reconstructed results, respectively. 
CD loss can be defined as  

\textcolor{black}{
\begin{equation}
\begin{split}
L_{CD}(S_i,S_o)=\frac{1}{2}(\frac{1}{|S_i|}\sum_{x \in S_i}\mathop{\min}_{y \in S_o}\|x-y\|_2\\
+\frac{1}{|S_o|}\sum_{x \in S_o}\mathop{\min}_{y \in S_i}\|x-y\|_2).
\end{split}
\end{equation}
}

We can see that CD measures the reconstruction loss through the average distance between points in $S_i$ or $S_o$ and their nearest neighbors in another point set.

Let $f_1(\cdot)$, $f_2(\cdot)$ be the combination of parameter-shared Multi Layer Perceptrons (MLPs) and symmetric pooling operations like PointNet~\cite{qi2017pointnet}, $Con(\cdot)$ be the concatenation of features, $g(\cdot)$ be a group of MLPs, \textbf{SiaCon} can be defined as
\begin{equation}
F_{io}=Con(f_1(S_i),f_1(S_o)).
%\begin{split}
%\mathcal{L}_{CD}(S_1,S_2)=\frac{1}{2}(\frac{1}{|S_i|}\sum_{x \in S_i}\mathop{\min}_{y \in S_o}\|x-y\|_2\\
%+\frac{1}{|S_o|}\sum_{x \in S_o}\mathop{\min}_{y \in S_i}\|x-y\|_2).
%\end{split}
\end{equation}

In \textbf{SiaAtt}, we have

\begin{equation}
\label{gio}
\left\{
\begin{split}
& F_i=g(Con(S_i,f_2(S_i),F_{io})),\\
& F_o=g(Con(S_o,f_2(S_o),F_{io})).
\end{split}
\right.
%seeds=s(X)
%C\!=\!seeds+MLPs(f(f(X),\!f(Y),\!seeds),\!seeds),\!
\end{equation}
The weight distributions can then be defined as 
%$ W_i=\frac{(\sigma+e^{-F_i^2})}{\sum (\sigma+ e^{-F_i^2})}$ and $W_o=\frac{(\sigma+e^{-F_o^2})}{\sum (\sigma+ e^{-F_o^2})}$,
\textcolor{black}{
\begin{equation}
\left\{
\begin{split}
& W_i=\frac{\sigma+e^{-F_i^2}}{|F_i| \cdot \sigma+\sum e^{-F_i^2}},\\
& W_o=\frac{\sigma+e^{-F_o^2}}{|F_o| \cdot \sigma+\sum e^{-F_o^2}},\\
\end{split}
\right.
\label{wio}
\end{equation}
}
where the boundary coefficient $\sigma$ is a small constant used to adjust the weight distribution.
\textcolor{black}{Intuitively speaking, in Eq.~\ref{wio}, $F_i$ is firstly scaled to $0 \sim 1$ with $e^{-F_i^2}$, where the scaled results will be normalized into weight distributions satisfying $\sum W_i =1$ and $\sum W_o=1$. $\sigma$ is used to soft the weight distributions and prevent each weight from being too small to optimize.}
% and prevent the denominator from 0.
The final loss measurement can be defined as

\textcolor{black}{\begin{equation}
\begin{split}
L_{R}(S_i,S_o)=\frac{1}{2}(\frac{1}{|S_i|}\sum_{x \in S_i} W_i \cdot\mathop{\min}_{y \in S_o}\|x-y\|_2
\\
+\frac{1}{|S_o|}\sum_{x \in S_o}W_o \cdot\mathop{\min}_{y \in S_i}\|x-y\|_2).
\end{split}
\end{equation}}

\textcolor{black}{Note that our method estimates the weight for each point in a same point cloud/sample, which is quite different with the boosting-related re-weighting methods to predict weights for various point clouds/samples in the dataset.}
%The main difference between boosting-related re-weighting methods and our method is that boosting methods should work on samples/point clouds, while our method estimates the weight for each point in a same point cloud/sample instead of the weights for point clouds in the dataset stage.

\subsection{Training Pipeline}
\label{training}
%In this section, we present the training procedure with LCD. 
LCD is trained with adversarial strategy to consistently search for existing shape differences between reconstructed results and input point clouds. 
\begin{algorithm}[t]
	\caption{Training Process with LCD.}
	\label{process}
	\begin{algorithmic}
		\STATE {\bfseries Input:} Input $S_i$, reconstructed results $S_o$, the number of iterations $iter$, the reconstruction network $RecNet(\cdot)$
		\FOR{$n=1$ {\bfseries to} $iter$}
		\STATE Calculate output of the reconstruction network:
		
		\STATE $S_o^n=RecNet(S_i^n)$.
		
		Let $\theta_{L}$ and $\theta_{R}$ be the parameters of LCD and the reconstruction network, respectively.
		\STATE Fix $\theta_{R}$ and optimize $\theta_{L}$ by descending gradient:
		
		$\nabla_{\theta_{L}} L_{LCD}(S_o^n,S_i^n)$.
		\STATE Fix $\theta_{L}$ and optimize $\theta_{R}$ by descending gradient:
		
		$\nabla_{\theta_{R}} L_R(S_o^n,S_i^n)$.
		\ENDFOR
	\end{algorithmic}
	\vskip -0.05in
\end{algorithm}

\textcolor{black}{The whole training process with LCD is a generative-adversarial process similar as GAN~\cite{NIPS2014_5ca3e9b1}, which updates the parameters of LCD and the reconstruction network by turns. In each iteration, LCD is optimized by $L_{LCD}$ to explore more shape differences, where the reconstruction network is then optimized with $L_R$ to eliminate the searched differences.}
Let $L_R$ be the reconstruction loss defined in Sec.~\ref{CAM}. We define the adversarial loss to optimize LCD as 
$L_{LCD}=-log(L_R+\sigma_r),$
where $\sigma_r$ is a tiny value to avoid errors when $L_R \rightarrow 0$.
%adversarial constraint and the reconstruction loss, respectively. The training process can be presented as Alg.~\ref{process}.
%Following PCLoss
\textcolor{black}{}

\section{Experiments}
\subsection{Dataset and Implementation Dtails}
\begin{table*}[ht]
	\normalsize
	\begin{center}
		\caption{Comparison with reconstruction losses. \textbf{Bold} marks the best results.}
		%\vskip -0.1in
		\label{reconsp}
		\scalebox{1.0}{
			\begin{tabular}{c|cccccccccccc}
				\toprule
				
				Networks& \multicolumn{2}{c|}{AE}                            & \multicolumn{2}{c|}{Folding}                       & \multicolumn{2}{c|}{AE(PN++)}                           & \multicolumn{2}{c|}{Folding(PN++)} & \multicolumn{2}{c|}{AE(DGCNN)}                           & \multicolumn{2}{c}{Folding(DGCNN)}
				\\ \midrule
				
				Metrics& MCD           & \multicolumn{1}{c|}{HD}            & MCD           & \multicolumn{1}{c|}{HD}            & MCD           & \multicolumn{1}{c|}{HD}            & MCD           & \multicolumn{1}{c|}{HD}
				& MCD           & \multicolumn{1}{c|}{HD}            & MCD           & HD           
				\\\midrule
				
				CD~\cite{fan2017point}     & 0.32          & \multicolumn{1}{c|}{1.87}          & 0.40          & \multicolumn{1}{c|}{4.13}          & 0.37          & \multicolumn{1}{c|}{2.50}          & 0.34          & \multicolumn{1}{c|}{3.37}   & 0.30          & \multicolumn{1}{c|}{1.88}            & 0.52           & 3.84       
				\\
				
				EMD~\cite{fan2017point}    & 0.25          & \multicolumn{1}{c|}{2.23}          & -             & \multicolumn{1}{c|}{-}             & 0.26          & \multicolumn{1}{c|}{2.51}          & -          & \multicolumn{1}{c|}{-}     & 0.21           & \multicolumn{1}{c|}{2.09}            & -           & -
				\\
				
				DCD~\cite{wu2021density}    & 0.28          & \multicolumn{1}{c|}{1.75}          & 0.91          & \multicolumn{1}{c|}{8.41}          &    0.28      &   \multicolumn{1}{c|}{1.84}          &     0.47      & \multicolumn{1}{c|}{5.43}       &     0.26       & \multicolumn{1}{c|}{1.86}            &      -      &       -    \\
				
				PCLoss~\cite{huang2022learning}   & 0.23          & \multicolumn{1}{c|}{1.66}          & 0.33          & \multicolumn{1}{c|}{2.57}          & 0.24          & \multicolumn{1}{c|}{1.87} & 0.31          & \multicolumn{1}{c|}{2.50}    & 0.20           & \multicolumn{1}{c|}{1.51}            & 0.43           & 3.10
				\\
				
				Ours & \textbf{0.22} & \multicolumn{1}{c|}{\textbf{1.51}} & \textbf{0.31} & \multicolumn{1}{c|}{\textbf{2.47}} & \textbf{0.24} & \multicolumn{1}{c|}{\textbf{1.66}}          & \textbf{0.28} & \multicolumn{1}{c|}{\textbf{2.37}}      & \textbf{0.20}          & \multicolumn{1}{c|}{\textbf{1.48}}            & \textbf{0.34}           & \textbf{2.45}
				\\ \bottomrule
			\end{tabular}
		}
	\end{center}
	\vskip -0.2in
	%	\vskip -0.1in 
\end{table*}
\textbf{Training details.}
%To train reconstruction networks, we use ShapeNet part dataset~\cite{achlioptas2018learning} containing 12288/1870/2874 models in the train/val/test splits. 
\textcolor{black}{ShapeNet part dataset~\cite{achlioptas2018learning} is composed of 12288/1870/2874 models in the train/val/test splits. For the reconstruction task, we train the networks on the train split of ShapeNet part dataset, while evaluating on its test split.
For the unsupervised classification, we still train networks on the train split of ShapeNet part dataset and use ModelNet10 and ModelNet40 containing 10 and 40 categories of CAD models to evaluate the classification accuracy following FoldingNet~\cite{yang2018foldingnet}.}
Each model consists of 2048 points randomly sampled from the surfaces of mesh models. In this work, learning rates of reconstruction networks and LCD are set as 0.0001 and 0.002, while $\sigma$ and $\sigma_r$ are set as 0.01 and 1e-8. The matching-based evaluation of CD~\cite{yuan2018pcn} is introduced to calculate the matching distances in LCD.

\textbf{Reconstruction Networks.} 
To compare LCD with existing reconstruction losses, we conduct comparisons based on multiple reconstruction networks.
%on the reconstruction performances and the unsupervised classification accuracy of 
% including multiple commonly-used point cloud reconstruction models. 
AE~\cite{achlioptas2018learning} and FoldingNet~\cite{yang2018foldingnet} are two classic and commonly used point cloud reconstruction networks, which have been used in many works \cite{rao2020global,huang2021rfnet,li2019pu,yuan2018pcn}.
In this work, we follow PCLoss~\cite{huang2022learning} to construct 6 reconstruction networks with three commonly-used encoders PointNet~\cite{qi2017pointnet}, PointNet++~\cite{qi2017pointnet++} and DGCNN~\cite{wang2019dynamic} and 2 basic decoders  AE~\cite{achlioptas2018learning} and FoldingNet~\cite{yang2018foldingnet}. 
The reconstruction performances of whole structures and unsupervised classification accuracy of intermediate representations are adopted to evaluate the performances of different training losses.
%In this work, We apply PointNet~\cite{qi2017pointnet}, PointNet++~\cite{qi2017pointnet++} and DGCNN~\cite{wang2019dynamic} to the encoder parts of AE~\cite{achlioptas2018learning} and FoldingNet~\cite{yang2018foldingnet} to construct diverse reconstruction networks. We use 128-dim bottleneck layers in the encoder parts following AE~\cite{achlioptas2018learning}.
%Besides, to build stronger reconstruction networks, we divide point clouds into multiple local regions following PointNet++~\cite{qi2017pointnet++} and apply AE and FoldingNet in each region to construct Local AE (LAE) and Local FoldingNet (LFolding) networks.
%Besides, we also apply AE~\cite{achlioptas2018learning} and FoldingNet~\cite{yang2018foldingnet} on multiple local regions from PointNet++~\cite{qi2017pointnet++} to construct stronger local AE(LAE) and local FoldingNet(LFolding) networks. 
%To make a fair evaluation, we retrain all networks under same settings with different training losses. 
\begin{figure}[t]
	%\vskip -0.2in
	\begin{center}
		\scalebox{1.0}{
			\centerline{\includegraphics[width=\linewidth]{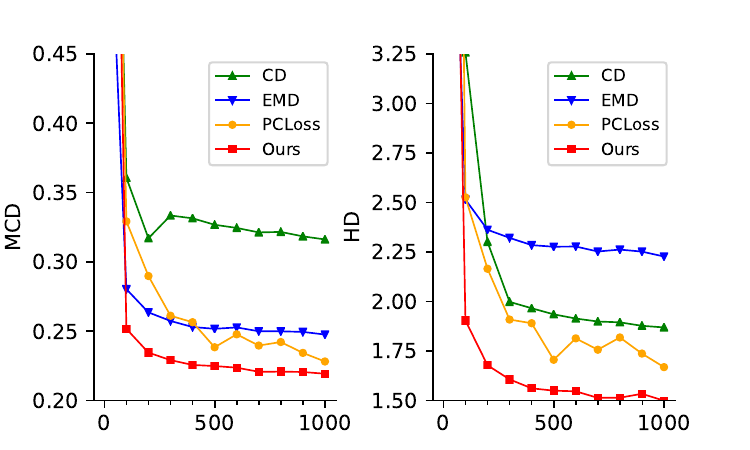}}
		}
		%\vskip -0.2in
		\caption{The error visualization during the whole training process.}
		%\vskip -0.2in
		\label{pic:tcurce}
	\end{center}
\end{figure}
\subsection{Comparison with Existing Reconstruction Loss}
%\textbf{Comparison on reconstruction errors.}
\textcolor{black}{To confirm the performance of our method, we follow PCLoss~\cite{huang2022learning} for the comparison settings. The reconstruction errors and performances of representation learning are adopted for evaluation.}
%In this section, we follow PCLoss~\cite{huang2022learning} for the comparison settings.
The qualitative and quantitative results are presented in Fig.~\ref{pic:quali} and  Table~\ref{reconsp}, respectively. Multi-scale Chamfer Distance (MCD) proposed by \cite{huang20193d,huang2022learning} and Hausdorff distance (HD) from \cite{wu2020pcc,huang2022learning} are used as metrics in this work.
%which are relatively robu
We can see that our method achieves lowest reconstruction errors on multiple reconstruction networks. 
As shown in the circled regions of Fig.~\ref{pic:quali}, our method can help the reconstruction network create clearer details such as the wings of airplane and the back of chairs, which confirms its effectiveness.
%It has clearer details and more complete shapes as shown in Fig.~\ref{pic:quali}, which confirms its effectiveness. 

%\textbf{Comparison on representation learning.}
The reconstruction networks can also be used to extract intermediate representations from point clouds for classification. We also conduct a comparison on unsupervised classification following AE~\cite{achlioptas2018learning}, FoldingNet~\cite{yang2018foldingnet}, and PCLoss~\cite{huang2022learning}. In details, the reconstruction networks are trained with different losses on ShapeNet~\cite{yi2016scalable} and adopted to extract representations from point clouds in ModelNet10 and ModelNet40~\cite{wu20153d}. The extracted representations will be used to train Supported Vector Machines (SVMs) with corresponding labels, where the classification accuracy can then reflect the distinguishability of representations. 

As shown in Table~\ref{mn_cls}, our method has higher classification accuracy than existing methods in most phenomena, which means LCD can help the reconstruction networks learn more representative representations.
\begin{table}[h]
	\normalsize
	%\setlength\tabcolsep{4pt}
	%\renewcommand\arraystretch{1.0}
	%	\vskip -0.1in
	%\caption{Comparison on unsupervised classification.}
	%\vskip -0.2in
	%\vskip -0.1in
	\caption{Comparison on unsupervised classification.}
	%\vskip -0.2in
	\label{mn_cls}
	\begin{center}
		\scalebox{1.0}{
			\begin{tabular}{c|c|cccc}
				\toprule
				\multirow{2}{*}{RecNet}        & \multirow{2}{*}{Dataset} & \multicolumn{4}{c}{Methods}                     \\
				&                          & CD    & EMD   & PCLoss           & Ours         \\ \midrule
				\multirow{2}{*}{AE}             & MN10                     & 90.60 & 89.49 & 91.48 & \textbf{91.48}          \\ %\cmidrule{2-6} 
				& MN40                     & 85.92 & 85.47 & 86.36          & \textbf{86.60} \\ \midrule
				\multirow{2}{*}{Folding}        & MN10                     & 91.03 & -     & \textbf{91.70} & 91.04         \\ %\cmidrule{2-6} 
				& MN40                     & 85.22 & -     & 85.35          & \textbf{86.81} \\ \midrule
				\multirow{2}{*}{\begin{tabular}[c]{@{}c@{}}AE\\ (PN++)\end{tabular}}       & MN10                     & 90.38 & 90.15 & 92.04          & \textbf{92.70} \\ %\cmidrule{2-6} 
				& MN40                     & 88.03 & 88.07 & 87.54          & \textbf{88.39} \\ \midrule
				\multirow{2}{*}{\begin{tabular}[c]{@{}c@{}}Folding\\ (PN++)\end{tabular}}  & MN10                     & 91.48 & -     & 91.48          & \textbf{92.59} \\ %\cmidrule{2-6} 
				& MN40                     & 87.01 & -     & 86.73          & \textbf{87.87} \\ \midrule
				\multirow{2}{*}{\begin{tabular}[c]{@{}c@{}}AE\\ (DGCNN)\end{tabular}}      & MN10                     & 91.37 & 91.26 & 92.37 & \textbf{92.81}         \\ %\cmidrule{2-6} 
				& MN40                     & 87.50 & 87.54 & \textbf{88.11} & 87.54          \\ \midrule
				\multirow{2}{*}{\begin{tabular}[c]{@{}c@{}}Folding\\ (DGCNN)\end{tabular}} & MN10                     & 91.26 & -     & 91.81          & \textbf{92.37} \\ %\cmidrule{2-6} 
				& MN40                     & 86.85 & -     & \textbf{87.50}& 86.65          \\ \midrule
				%				\multirow{2}{*}{GLRNet}         & MN10                     & 93.58 & -     & -              & \textbf{95.24} \\ %\cmidrule{2-6} 
				%				& MN40                     & 91.07 & -     & -              & \textbf{91.31} \\ 
				\bottomrule
			\end{tabular}
		}
	\end{center}
	%\vskip -0.3in
\end{table}

\subsection{Training process analysis}
To analyze the training process when optimizing the reconstruction networks with LCD. We visualize and compare the reconstruction errors during the iterations between our method and a few representative training losses including CD~\cite{yuan2018pcn}, EMD~\cite{fan2017point}, PCLoss~\cite{huang2022learning} based on AE~\cite{achlioptas2018learning}. The results are presented in Fig.~\ref{pic:tcurce}. We can see that LCD has much faster and steadier convergence than existing methods. Beside, it performs much better than totally learning-based PCLoss at 0 $\sim$ 200 iterations, which confirms that the introducing of static matching-based evaluation can ensure the performances at the beginning of training process.

\subsection{Comparison on Training Efficiency}
%Training efficiency is another metric worth consideration for reconstruction losses. 
In this section, we compare the time cost consumed by a single iteration between different methods. The results are presented in Table~\ref{comeff}. Although LCD is slower than CD and DCD, it has better performances as shown in Table~\ref{reconsp}. We can see that LCD has higher efficiency than the totally learning-based reconstruction loss PCLoss due to its more concise designation, which can further confirm its effectiveness.
\begin{table}[th]
	\normalsize
	%\vskip -0.1in
	%\setlength\tabcolsep{6pt}
	%\begingroup
	%\setlength{\tabcolsep}{3.0pt} % Default value: 6pt
	\caption{Training efficiency comparison conducted on an NVIDIA 2080ti with a 2.9GHz i5-9400 CPU.}
	%\vskip -0.2in
	\label{comeff}
	\begin{center}
		\scalebox{1.0}{
			\begin{tabular}{c|ccc|cc}
				\toprule
				& \multicolumn{3}{c|}{Non-learning} & \multicolumn{2}{c}{Learning-based}      \\ \midrule
				Methods  & CD                  & EMD         & DCD  & PCLoss & Ours        \\ \midrule
				Time(ms) & \textbf{23}         & 216         & 23    & 57     & \textbf{43} \\ \bottomrule
			\end{tabular}
		}
	\end{center}
	%\vskip -0.3in
	%\endgroup
\end{table}
\begin{figure}[t]
	%\vskip -0.2in
	\begin{center}
		\scalebox{1.0}{
			\centerline{\includegraphics[width=\linewidth]{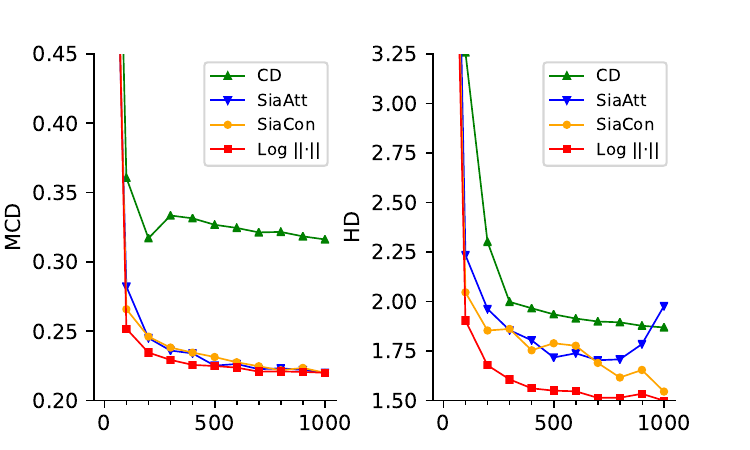}}
		}
		%\vskip -0.1in
		\caption{\textcolor{black}{Ablation study for components over iterations.}}
		\label{pic:abcurve}
		%\vskip -0.2in	
	\end{center}
\end{figure}
\begin{figure}[t]
	%\vskip -0.3in
	\begin{center}
		\scalebox{1.0}{
			\centerline{\includegraphics[width=\linewidth]{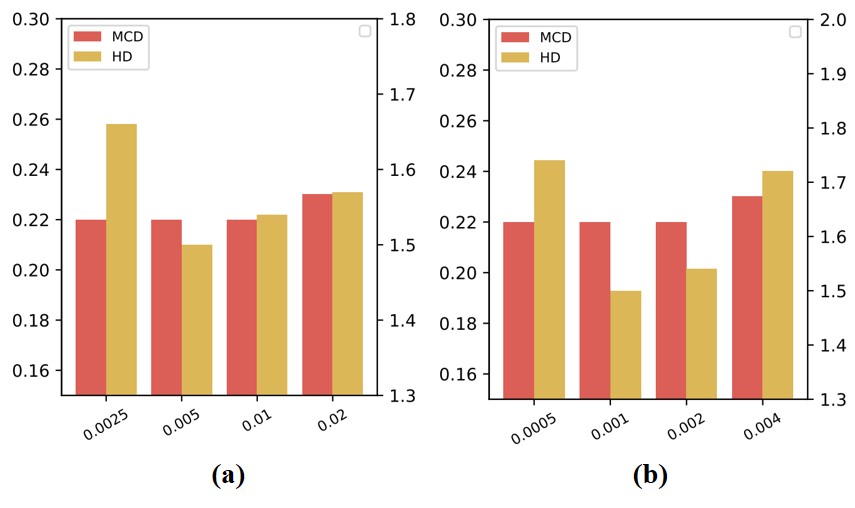}}
		}
		%\vskip -0.1in
		\caption{Influences of $\sigma$ (a) and LCD learning rate (b).}
		%\vskip -0.3in
		\label{pic:aba}
	\end{center}
\end{figure}
\subsection{Ablation Study}

\textbf{Ablation study for the components.}
In this section, we explore the effect of proposed components by removing them and retraining the networks. \textbf{SiaAtt} and \textbf{SiaCon} denote the Siamese Attention block and Siamese Concatenation block, respectively. $log\|\cdot\|$ means the $log\|\cdot\|$ operation mentioned in Sec.~\ref{training} to dynamically adjust the optimization of LCD.
The results are presented in Table~\ref{comab}. We can see that removing any component would reduce the final performance. 
\textcolor{black}{To compare the effect of these components more intuitively, we also visualize the reconstruction errors over all the iterations in Fig.~\ref{pic:abcurve}. We can see that with adding the \textbf{SiaAtt}, \textbf{SiaCon}, and $log\|\cdot\|$ gradually makes the errors reduce fast and stably over the whole iterations.}
An interesting condition is that \textbf{SiaAtt} reduces MCD while slightly increasing the HD metric at the end of iterations. It may come from the lack of perception for the overall input and output shapes, making it difficult to find the regions with larger reconstruction errors and mislead the training of reconstruction networks. This condition is then addressed by injecting overall shape features with \textbf{SiaCon}.
\begin{table}[th]
	\normalsize
	%\vskip -0.1in
	%\setlength\tabcolsep{3.5pt}
	\begingroup
	\caption{Ablation for components.}
	%\vskip -0.2in
	\label{comab}
	\begin{center}
		\scalebox{1.0}{
			\begin{tabular}{cccc|cc}
				\toprule
				CD & \textbf{SiaAtt} & \textbf{SiaCon} &  $log\|\cdot\|$   & MCD & HD \\ \midrule
				\checkmark&     &         & &  0.32   &  1.87   \\
				\checkmark&  \checkmark   &     &    &  0.22   &   1.98  \\
				\checkmark&   \checkmark  &   \checkmark    &    &  0.22    & 1.54  \\ 
				\checkmark&   \checkmark  &   \checkmark    &  \checkmark  &   \textbf{0.22}  &   \textbf{1.51} \\
				\bottomrule
			\end{tabular}
		}
	\end{center}
	%\vskip -0.2in
	\endgroup
\end{table}

\textbf{Influence of the boundary coefficient $\sigma$.}
The boundary coefficient $\sigma$ defined in Sec.~\ref{CAM} may  affect the weight distribution for matching distances. Here, we present experiments to explore its influence as shown in Fig.~\ref{pic:aba}-a. We can see that larger or smaller $\sigma$ both have negative influences on the results. According to Eq.~\ref{wio}, too small $\sigma$ makes the distribution steep and hard to train, while too big $\sigma$ may cause the distribution over-smoothing and limit its performance.

\textbf{Influence of the LCD learning rate.}
The LCD learning rate decides the convergence  and has influence on final performance. We conduct a group of experiments to observe the influence of the LCD learning rate. The results are presented in Fig.~\ref{pic:aba}-b. We can see that too small or large learning rates both reduce performances. Small learning rates may limit the ability of LCD to search shape differences, while larger learning rates may lead to its unsteady convergence.

\section{Conclusion}
In this work, we propose a simple but effective point cloud reconstruction loss, named Learnable Chamfer Distance (LCD), by combining the dynamic learning-based strategy and static matching-based evaluation in a more reasonable way. 
LCD dynamically predicts weight distributions for matching distances of different points, which is optimized with adversarial strategy to search and pay more attention to regions with larger shape defects. Benefited from the reasonable combination of matching-based evaluation and learning-based strategy, LCD has both faster convergence and higher training efficiency than totally learning-based PCLoss.
According to the experiments on multiple reconstruction networks, LCD can help the reconstruction networks achieve better reconstruction performances and extract more representative representations.

\section*{Acknowledgment}
We thank all reviewers and the editor for excellent contributions. 
This work is supported by the Key Scientific and Technological Innovation Project of Hangzhou under Grant 2022AIZD0019.
%Though existing learning-based algorithms have achieved satisfied task-oriented performances by optimizing through task networks, they are strictly limited by the differentiability of task network. Specifically speaking, existing learning-based pipelines are not appropriate for task networks non-differentiable to input point clouds such as PointNet++~\cite{qi2017pointnet++} which adopts the non-differentiable KNN operation to aggregate features. We will focus on adapting the learning-based sampling strategies to more diverse non-differentiable task networks in the future.

\bigskip
%%

%\newpage
{\small
	\bibliographystyle{elsarticle-num}
	\bibliography{egbib}
}






% \end{thebibliography}

\end{document}